\documentclass[letterpaper]{article} 
\usepackage{aaai2026}  
\usepackage{times}  
\usepackage{helvet}  
\usepackage{courier}  
\usepackage[hyphens]{url}  
\usepackage{graphicx} 
\usepackage{natbib}  
\usepackage{caption} 
\usepackage{algorithm}

\usepackage{adjustbox}
\usepackage{amsmath}
\usepackage{amssymb}
\usepackage{amsthm}
\usepackage{booktabs}
\usepackage{multirow}
\usepackage{tabularx}
\usepackage{subcaption}
\usepackage{algpseudocode}

\newcommand{\eg}{\emph{e.g.}}
\newcommand{\ie}{\emph{i.e.}}

\newcommand{\etal}{\emph{et al.}}

\usepackage{newfloat}
\usepackage{listings}
\DeclareCaptionStyle{ruled}{labelfont=normalfont,labelsep=colon,strut=off} 
\floatstyle{ruled}
\newfloat{listing}{tb}{lst}{}
\floatname{listing}{Listing}
\title{Test-Time Preference Optimization for Image Restoration}
\author {
    Bingchen Li\textsuperscript{\rm 1}\textsuperscript{\dag},
    Xin Li\textsuperscript{\rm 1}\thanks{Corresponding Author.},
    Jiaqi Xu\textsuperscript{\rm 2},
    Jiaming Guo\textsuperscript{\rm 2},
    Wenbo Li\textsuperscript{\rm 3},
    Renjing Pei\textsuperscript{\rm 2},
    Zhibo Chen\textsuperscript{\rm 1}
}
\affiliations {
    \textsuperscript{\rm 1}University of Science and Technology of China,\\
    \textsuperscript{\rm 2}Huawei Noah's Ark Lab,\\
    \textsuperscript{\rm 3}Joy Future Academy\\
    lbc31415926@mail.ustc.edu.cn, xin.li@ustc.edu.cn,
    jiaqixuac@gmail.com
}

\begin{document}

\maketitle

\renewcommand\thefootnote{\dag}
\footnotetext{This work was done during an internship at Huawei Noah's Ark Lab.}

\begin{abstract}

Image restoration (IR) models are typically trained to recover high-quality images using L1 or LPIPS loss.
To handle diverse unknown degradations, zero-shot IR methods have also been introduced. However, existing pre-trained and zero-shot IR approaches often fail to align with human preferences, resulting in restored images that may not be favored.
This highlights the critical need to enhance restoration quality and adapt flexibly to various image restoration tasks or backbones without requiring model retraining and ideally without labor-intensive preference data collection.
In this paper, we propose the first Test-Time Preference Optimization (TTPO) paradigm for image restoration, which enhances perceptual quality, generates preference data on-the-fly, and is compatible with any IR model backbone.
Specifically, we design a training-free, three-stage pipeline:
(i) generate candidate preference images online using diffusion inversion and denoising based on the initially restored image;
(ii) select preferred and dispreferred images using automated preference-aligned metrics or human feedback;
and (iii) use the selected preference images as reward signals to guide the diffusion denoising process, optimizing the restored image to better align with human preferences.
Extensive experiments across various image restoration tasks and models demonstrate the effectiveness and flexibility of the proposed pipeline.

\end{abstract}


\section{Introduction}

Image restoration (IR) aims to remove unpleasant distortions from real-world low-quality (LQ) images and recover their high-quality (HQ) counterparts.
Typical IR tasks include
image denoising \cite{denoise5},
image deblurring \cite{deblur1,deblur5},
image super-resolution \cite{li2024ucip,li2024sedgan},
weather artifact removal \cite{derain2,li2018benchmarking},
and all-in-one restoration \cite{airnet}.
Despite the variety of IR models and advancements in zero-shot IR approaches \cite{kawar2022denoisingddrm} to address unknown degradations, they often struggle to align with human preferences.
For example, image restoration models optimized with standard L1 loss may achieve high PSNR scores but produce over-smoothed outputs that are less favored by users, highlighting the need for preference-based optimization.

Increasing alignment between model outputs and human preferences has been actively explored in tasks like text-to-image (T2I) generation~\cite{t2isurvey,t2isurvey2}.
Reinforcement learning (RL) is often used for this purpose by fine-tuning T2I models, leveraging human-annotated preference data, (\eg, Pick-a-Pic~\cite{pickapic}) to train reward models \cite{ouyang2022trainingrlhf} or directly guide model fine-tuning \cite{wallace2024diffusiondpo}.
While similar techniques could be applied to IR tasks, several challenges arise.
First, existing image restoration models feature diverse architectural designs, with many being deterministic during inference, which makes them theoretically unsuitable as policy models under modern RL frameworks (\eg, PPO \cite{schulman2017proximalppo}, DPO \cite{rafailov2023directdpo}).
Furthermore, there are no publicly available preference datasets specifically tailored for IR tasks, as collecting such datasets with human annotations is both labor-intensive and challenging.

\begin{figure}
    \centering
    \includegraphics[width=\linewidth]{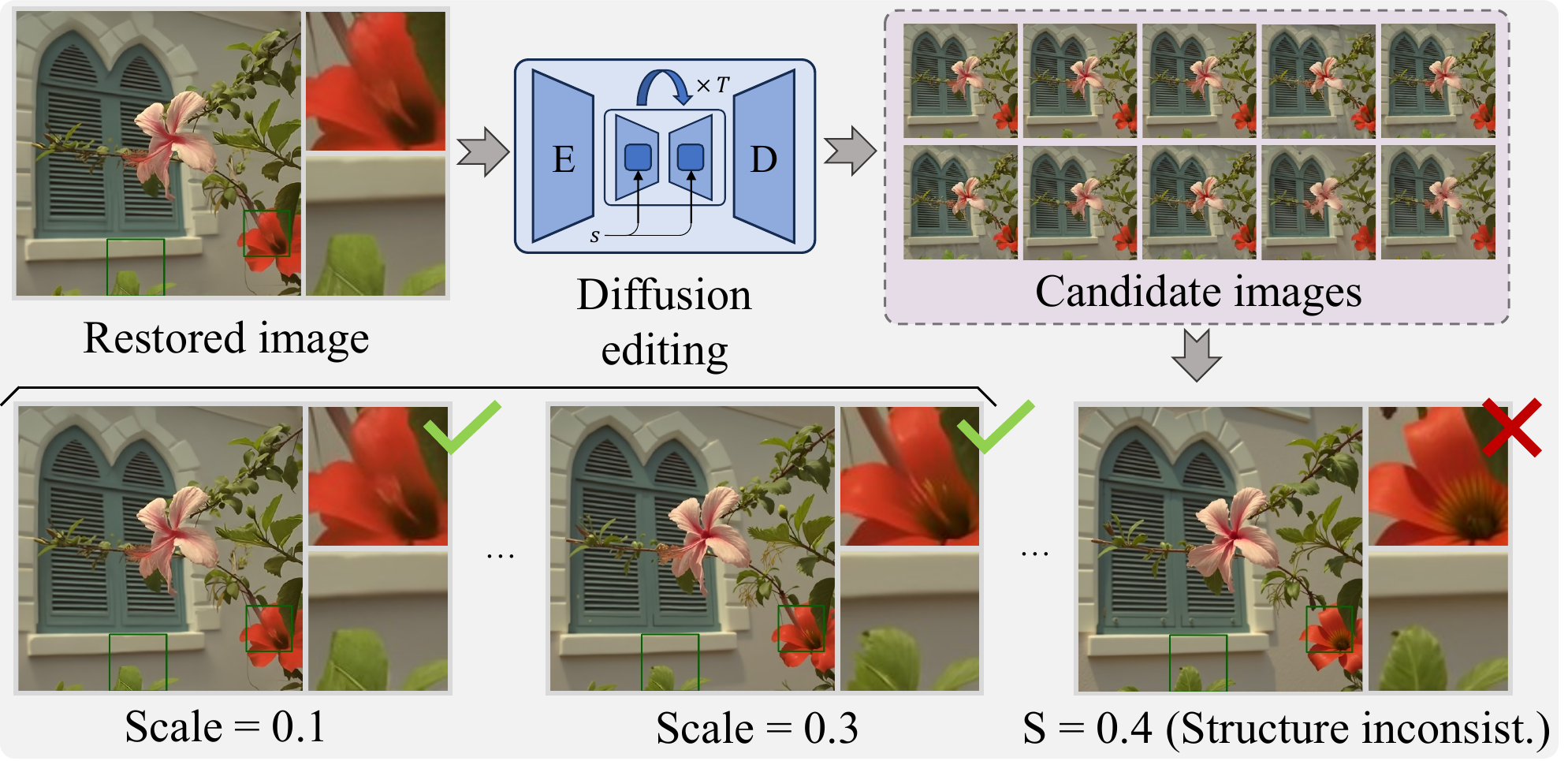}
    \caption{
    We use diffusion editing to generate candidate preference images.
    To preserve structural consistency, the noise scale is limited to a suitable range.
    }
    \label{fig:intro}
\end{figure}

In this work, we propose Test-Time Preference Optimization (TTPO), a flexible approach to enhance restoration results for optimized perceptual quality, compatible with a wide range of IR tasks and backbones.
The entire pipeline is broadly divided into three key stages: generation, selection, and optimization.
In the first generation stage, multiple candidate preference images for restoration are generated on-the-fly for each image using pre-trained diffusion models.
This is achieved by leveraging diffusion inversion \cite{meng2022sdedit} to modify the initially restored images while adjusting the scale of the added noise.
Notably, using a large noise scale can lead to structural inconsistencies, as shown in Fig.~\ref{fig:intro} (structural inconsistency), which is critical for IR tasks.
To address this, we carefully limit the noise scale to an appropriate range when generating preference images.
Furthermore, we ensemble multiple diffusion models in this process to enrich the diversity of candidate preference images.
Note that this preference image generation process is model-agnostic and applicable to the initially restored images from any IR backbone.

In the second stage, selection, we identify preferred and dispreferred images from the generated restoration samples.
While this process could ideally involve manual selection by humans to reflect true preferences, such involvement becomes impractical for large-scale datasets.
To address this, we opt to use automated evaluation metrics to approximate human preferences.
Specifically, no-reference image quality assessment (NR-IQA) metrics~\cite{wu2023q,you2025teachingdeqa,wu2025visualquality} are well-suited for evaluating restored image quality, as ground truth images are typically unavailable during testing.
To improve selection quality, we adopt a metric combination strategy that integrates top human-aligned metrics together to rank images, as individual metrics may deviate from human preferences.

In the final optimization stage, we refine the initially restored image using conditioned diffusion denoising, guided by the selected preference samples.
This controlled optimization process enhances image details while incorporating preference alignment.
Specifically, at each reverse denoising step, the clean image is firstly predicted by a pre-trained diffusion pipeline~\cite{ddpm}.
To guide the actual subsequent denoising step, we modulate the process using preference samples by calculating the distance between the predicted clean image and the preferred/dispreferred samples.
These distance measurements serve as reward signals, generating gradient descent directions that condition the denoising process, steering it toward the preferred direction while avoiding the dispreferred one.
Additionally, the initially restored image is integrated as a constraint to preserve the overall image structure throughout the optimization.
Various techniques, including frequency-based guidance decomposition and stage-wise guidance, are designed to enhance the optimization process.

Overall, the contributions of this paper can be summarized as follows:
\begin{itemize}
    \item To the best of our knowledge, we introduce the first test-time preference optimization paradigm for image restoration, requiring neither model retraining nor extensive human preference data collection. Instead, the proposed pipeline can be seamlessly integrated with any existing IR model and directly optimizes restored images.
    \item We present an innovative approach that leverages diffusion-based editing to generate preference samples on-the-fly and employs conditioned diffusion denoising for preference optimization. Furthermore, our pipeline offers high flexibility by allowing human involvement in the selection of preference images.
    \item We conduct comprehensive experiments, including 12 quantitative comparisons and a user study, demonstrating the versatility of our proposed pipeline across a wide range of IR tasks and models, achieving improvements in both objective metrics and subjective human evaluations.
    
\end{itemize}

\section{Related Works}

\subsection{Image Restoration Models}

Image restoration (IR) has witnessed the development of diverse model architectures, ranging from traditional CNN-based~\cite{srcnn} to recent Transformer-based~\cite{HAT,quan2024siamese}, diffusion-based~\cite{li2023diffusionsurvey,ren2025moe,qincamedit}, MLP-based~\cite{tu2022maxim,li2024ucip}, and state-space model-based~\cite{guo2025mambairv2} designs, reflecting the evolving needs for a better restoration fidelity and perceptual quality. Despite advances in network designs, the training objective for most IR models remains largely unchanged. Early works~\cite{srcnn} primarily focus on optimizing pixel-wise fidelity using L1/MSE losses. However, such objectives often lead to over-smoothed outputs. To address this limitation, some subsequent approaches~\cite{bsrgan,realesrgan} incorporate generative adversarial networks (GANs) into the training pipeline to produce more perceptually pleasing results. While numerous network designs have emerged for IR, little attention has been given to preference optimization, leading to restored images that often misalign with human perceptual preferences.

\subsection{Zero-Shot Diffusion-Based Image Restoration}
Among various IR models, there is a special type of diffusion-based IR model worth noting. Zero-shot diffusion-based IR models aim to leverage the strong generative prior in the pre-trained diffusion model (\eg, DDPM~\cite{ddpm}) to restore images. Song \etal~\cite{song2022solving} propose to use gradient guidance to address inverse problems in medical imaging. ILVR~\cite{choi2021ilvr} incorporates low-frequency condition information from the reference image as guidance. DDRM~\cite{kawar2022denoisingddrm} proposes an SVD-based decomposition solution to improve guidance information. DDNM~\cite{wang2023zeroddnm} further enhances the solution with range-null space decomposition. DPS~\cite{chung2023diffusiondps} predicts noise-clean images at each denoising step, obtaining gradient guidance for the next step denoising through distance measurements. GDP~\cite{fei2023generativegdp} shares a similar idea, while TAO~\cite{gou2024testTAO} further improves the distance measurements by incorporating GAN loss.

\subsection{Diffusion Preference Optimization}

Diffusion preference optimization typically aims to align pre-trained T2I generation models to generate more visually appealing results. Diffusion-DPO~\cite{wallace2024diffusiondpo} introduces DPO~\cite{rafailov2023directdpo} into T2I generation and fine-tunes diffusion models with pairwise preference data to achieve better prompt alignment and aesthetic scores. MaPO~\cite{mapo} handles reference mismatch by replacing divergence regularization on the reference model with an amplification factor. Some recent works~\cite{lee2025calibrated,gu2024diffusionrpo,yang2024densepo} focus on reward design and further improve the alignments of T2I models. To improve the training process of diffusion-DPO, Curriculum DPO~\cite{croitoru2025curriculumdpo} proposes a curriculum learning approach, gradually introducing harder preference pairs to improve model alignment performance.

\section{Preliminaries}

\subsection{Denoising Process}

The diffusion process is the key operation in diffusion models, where the objective is to remove noise from the noisy data samples \(x_t\) to recover the clean data \(x_0\), or conversely, to add noise in the forward process. Specifically, we focus on the Euler flow-matching scheduler employed in state-of-the-art diffusion models~\cite{esser2024scalingsd3,flux}:
\begin{equation}
    \label{equ:1}
    x_{t'} = x_t + \Delta t \cdot v_\theta(x_t,t),
\end{equation}
where each timestep \(t\in [0,1], \Delta t=t-t'\) is the interval between two timesteps, \(t', v_\theta(x_t,t)\) denotes the next sampled timestep and the learned velocity field, respectively. Notably, the clean data \(x_0\) can be predicted by Eq.~\ref{equ:1} by setting \(t'\) to 0: \(x_0 = x_t + t \cdot v_\theta(x_t,t)\). Conversely, the intermediate noisy latent \(x_t\) cin the forward process can be sampled using Eq. 13 from~\cite{esser2024scalingsd3}:
\begin{equation}
    \label{equ:2}
    x_t=(1-t)x_0+t\epsilon,
\end{equation}
where \(\epsilon\sim\mathcal{N}(0,I)\) denotes standard Gaussian noise.

\subsection{Preference Optimization}

Preference optimization is typically achieved by training models on preference data triplets \((c,x_w,x_l)\), where \(x_w\) and \(x_l\) represent the preferred (win) and dispreferred (lose) samples, and \(c\) denotes the condition used to generate these samples.
The probability of such preferences is often formulated by Bradley-Terry model~\cite{btmodel}: 
\begin{equation}
    P(x_w \succ x_l\mid c) = \sigma(r(x_w,c)-r(x_l,c)),
\end{equation}
where \(\sigma(\cdot)\) is the sigmoid function and \(r\) denotes the reward model used to represent human preferences. Due to the challenges of learning such a reward model, Rafailov~\etal~\cite{rafailov2023directdpo} propose directly optimizing the model using preference data \( (c,x_w,x_l) \sim \mathcal{D}\):
\begin{multline}
    \mathcal{L}_{\text{DPO}} =
    -\mathbb{E}_{(c, x_w, x_l) \sim \mathcal{D}} \biggl[ \log \sigma \biggl( \beta \cdot \log \frac{p_\theta(x_{w}\mid c)}{p_{\text{ref}}(x_{w}\mid c)} \\
    - \beta \cdot \log \frac{p_\theta(x_{l}\mid c)}{p_{\text{ref}}(x_{l}\mid c)} \biggr) \biggr],
\end{multline}
where \(p_{\text{ref}}, p_\theta\) denote the fixed pre-trained model and the fine-tuned model to align the preference, respectively, and \(\beta\) controls the regularization strength.

\section{Methods}

\begin{figure*}
    \centering
    \includegraphics[width=\hsize]{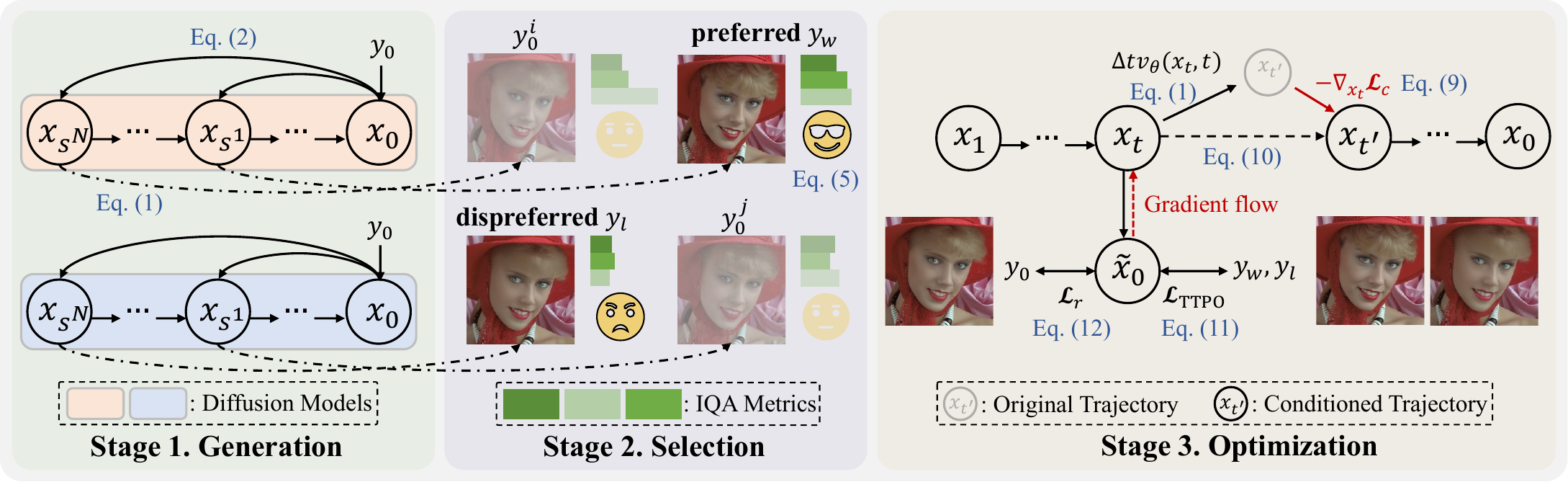}
    \caption{The pipeline of TTPO, which follows a generation-selection-optimization paradigm to perform test-time preference optimization for restored images. The pseudo codes are provided as Algorithm 1 in Sec.~\ref{sec:algo}.}
    \label{fig:method}
\end{figure*}

We propose the training-free test-time preference optimization (TTPO) approach for image restoration, which dynamically generates preference samples online and leverages them as reward signals to optimize the initially restored image from any image restoration backbone. This enhances visual preferences through a conditioned diffusion denoising process.
TTPO consists of three stages: (i) generation of candidate preference images in Sec.~\ref{sec:stage1}; (ii) selection of preference images in Sec.~\ref{sec:stage2}; and (iii) optimization of restored images in Sec.~\ref{sec:stage3}.

\subsection{Generation of Candidate Preference Images}
\label{sec:stage1}

In this stage, candidate preference images are generated for a given initially restored image from any image restoration backbone.
The success of preference optimization often relies on human-annotated preference datasets \cite{pickapic}.
However, for image restoration (IR) tasks, such preference datasets are unavailable, and collecting preference data is highly labor-intensive. Therefore, we propose generating preference data $\mathcal{D}$ at test time in TTPO. Leveraging the powerful editing capabilities of pre-trained diffusion models, this stage employs diffusion inversion \cite{meng2022sdedit} to generate diverse preference samples.

Specifically, let \(y_0\) represent the initially restored image from an IR model.
First, we obtain its noisy latent \( x_s = (1-s) x_0 + s \epsilon \) using Eq.~\ref{equ:2}, where \(s\in[0,1]\) controls the scale of the added noise, and $x_0$ is the encoded latent via VAE encoder $\mathcal{E}$.
Next, the edited version \(y_0^i\) is generated by progressively removing noise from \(x_s\) using diffusion denoising (Eq.~\ref{equ:1}) and decoding the latent back into an image in the end.
Notably, the inference of pre-trained diffusion models implicitly incorporates rich generative priors, such as texture and detail generation, into the edited image. This ensures the diversity of the generated preference images.
However, we observe that a large scale \(s\), such as $s=0.4$, leads to significant structural alternation (Fig.~\ref{fig:intro}). To preserve structural consistency, we limit \(s\) to the range $[0.1, 0.3]$.
Using this approach for a single sample, we construct the preference image set \(\mathcal{D}=\{y_0, y_0^1,y_0^2,\dots,y_0^N\}\) by leveraging multiple pre-trained diffusion models and varying \(s\), where \(N\) is the number of generated preference samples.

\subsection{Selection of Preference Images}
\label{sec:stage2}
After generating the preference samples, our goal is to select the preferred (win) sample \(y_w\) and the dispreferred (lose) sample \(y_l\).
While human involvement offers the most reliable indication of preference, it is labor-intensive and often impractical for large-scale applications.
To overcome this challenge, we propose a simple yet effective alternative by utilizing no-reference image quality assessment (NR-IQA) metrics as a proxy for preference selection.

However, relying on a single NR-IQA metric may not always align well with human perception \cite{zhang2022perceptualnriqa}. To address this, we adopt a hybrid evaluation approach by combining multiple metrics, consistent with findings in \cite{chen2024restoreagent,lin2025jarvisir}.
To implement this, we conduct a subjective experiment (details provided in Sec.~\ref{sec:nriqa}) to identify the top three metrics from the $\texttt{pyiqa}$ package \cite{pyiqa} that best align with human preferences. Using these selected metrics, we construct the hybrid metric by assigning a reward to each sample \(y_0^i\sim \mathcal{D}\) based on the average of the Z-score normalized values from the three NR-IQA metrics:
\begin{equation}
r_{y_0^i} = \bigl( \text{IQA}_1 + \text{IQA}_2 + \text{IQA}_3 \bigr) \ / \ 3.
\end{equation}
These rewards are then used to rank the candidate samples, with the top-ranked image selected as \(y_w\) and the bottom-ranked image as \(y_l\).

\begin{table*}[t]
\centering
\begin{adjustbox}{width=\textwidth}
\setlength{\tabcolsep}{8pt}
\begin{tabular}{@{}lcccccccccccc@{}}
\toprule

        & $y_0$ & $y_{\text{TTPO}}$ & $y_w$ & $y_l$ & $y_0$ & $y_{\text{TTPO}}$ & $y_w$ & $y_l$ & $y_0$ & $y_{\text{TTPO}}$ & $y_w$ & $y_l$ \\
        \cmidrule(lr){2-5}  \cmidrule(lr){6-9} \cmidrule(lr){10-13}

& \multicolumn{4}{c}{(a) DN (Kodak) - }    & \multicolumn{4}{c}{(b) DN (Kodak) - } & \multicolumn{4}{c}{(c) DN (Kodak) - }  \\

& \multicolumn{4}{c}{SwinIR~\cite{liang2021swinir}}    & \multicolumn{4}{c}{Restormer~\cite{zamir2022restormer}} & \multicolumn{4}{c}{DMID~\cite{dmiddenoising}}  \\

\midrule

MUSIQ   & 73.790 & \textbf{74.242} & 73.816 & 70.210 & 74.059 & \underline{74.105} & 74.280 & 70.401 & 73.999 & 73.893 & 74.249 & 70.175 \\

MANIQA  & 0.4565 &  \textbf{0.4912}  &  0.4828  &  0.4317  & 0.4655 &         \textbf{0.4928} &  0.4858 & 0.4361 & 0.4664 & \textbf{0.4832}  & 0.4765 & 0.4276\\

Q-Align &  3.9830 &  \underline{4.3201}  & 4.3662  & 3.9122  &  4.5054 &         \underline{4.6056} & 4.6706 & 4.3326 &  4.5736 & \underline{4.6276} & 4.6696 & 4.3391\\
NIQE $\downarrow$    &  4.1567 &  4.3001  &  4.1244 & 3.8924 &  4.2216 &  \textbf{3.9546}      &  3.9956 & 3.9148 & 4.1986 & \textbf{4.0079} & 4.0385 &  3.8991  \\
CLIPIQA & 0.5965 &  \underline{0.6117} & 0.6333 & 0.6342  & 0.5937 &  \underline{0.6015} & 0.6198 & 0.5938 & 0.6147 & 0.6097 & 0.6221 & 0.6030 \\
LIQE    &  4.2576 &  \textbf{4.8690}  &  4.8051 &  4.6102  & 4.2599 & \underline{4.8121}       & 4.8233 & 4.4915 & 4.4391 & \textbf{4.8450} &  4.8004 & 4.5659       \\
\midrule \midrule

& \multicolumn{4}{c}{(d) CISR (BSD100) - }    & \multicolumn{4}{c}{(e) CISR (BSD100) - } & \multicolumn{4}{c}{(f) RISR (RealSR) - }  \\ 

& \multicolumn{4}{c}{SwinIR~\cite{liang2021swinir}}    & \multicolumn{4}{c}{HAT~\cite{HAT}} & \multicolumn{4}{c}{StableSR~\cite{stablesr}}  \\

\midrule

MUSIQ   & 57.720 & \underline{57.764} & 59.540 & 50.249 & 57.798 & \underline{58.066} & 59.588 & 50.785 & 65.883 & \underline{66.096} & 66.694 & 62.193 \\

MANIQA
& 0.3499 & \textbf{0.3745}  &  0.3691 & 0.2811 & 0.3510 & \textbf{0.3778} & 0.3702     & 0.2828 & 0.4275 & \underline{0.4474} & 0.4488 & 0.3852 \\

Q-Align & 2.9806 & \underline{3.0110} & 3.1874  & 2.6024  & 2.9831       & \underline{3.0291}  & 3.1988 & 2.6006 & 3.2766 & \underline{3.3900} &  3.4917  & 3.1204 \\
NIQE $\downarrow$    & 6.1102 & 6.2586 & 5.8336 & 5.9172  & 6.0029  &  6.1579 & 5.7606 & 5.8992 & 5.8812 & \underline{5.7778} & 5.4205  &  5.3647 \\
CLIPIQA &  0.5300 & 0.5284 &  0.5518 & 0.4397 & 0.5399       & 0.5328 & 0.5555 &  0.4456 & 0.6233 & 0.5518 & 0.5979 & 0.5382 \\
LIQE    & 3.0100 &  \underline{3.0819} & 3.2359 & 2.7823  & 3.0343       & \underline{3.1246} &  3.2639 & 2.8359 & 3.5831 &  \underline{3.6191} & 3.7920 & 3.5152 \\
\midrule \midrule

& \multicolumn{4}{c}{(g) RISR (RealSR) - }    & \multicolumn{4}{c}{(h) DR (Rain100L) - } & \multicolumn{4}{c}{(i) DR (Rain100L) - }  \\ 

& \multicolumn{4}{c}{OSEDiff~\cite{wu2024oneosediff}}    & \multicolumn{4}{c}{Restormer~\cite{zamir2022restormer}} & \multicolumn{4}{c}{PromptIR~\cite{potlapalli2023promptir}}  \\

\midrule

MUSIQ   & 69.089 & \underline{69.138} & 69.723 & 65.887 & 71.558 & \underline{72.181} & 72.802 & 69.099 & 70.555 & \underline{71.413} & 72.214 & 67.994 \\

MANIQA  & 0.4717 & \textbf{0.4837} & 0.4816 &  0.4254 & 0.5200 & \textbf{0.5424} & 0.5422 & 0.4718 & 0.5063 & \textbf{0.5478} &  0.5437 & 0.4678 \\

Q-Align & 3.6962 & \underline{3.7458} & 3.8086 &  3.4310 & 4.0219 & \textbf{4.2553} &  4.2325 & 3.7953 & 3.8595 &  \underline{4.0759} & 4.1588 & 3.6656 \\
NIQE $\downarrow$    &  5.6474 & 5.8434  & 5.4493 & 5.4818  & 3.7239 & \underline{3.6696} & 3.6081 & 3.4614 & 3.1646 & 3.6794 & 3.4022  & 3.3319 \\
CLIPIQA & 0.6693  & 0.5875 & 0.6428 & 0.5724 &  0.7277 & \underline{0.7331} &  0.7440 & 0.7001 & 0.7691 & 0.7286 & 0.7623  & 0.7190 \\
LIQE    & 4.1216 & 4.0495 & 4.1909  & 3.9410  & 4.7392 &  4.7255 & 4.7240 & 4.1047  & 4.7529 &  \textbf{4.7820} & 4.7167 & 4.0718 \\
\midrule \midrule

& \multicolumn{4}{c}{(j) LLIE (CDD11) - }    & \multicolumn{4}{c}{(k) LLIE (CDD11) - WeatherDiff} & \multicolumn{4}{c}{(l) LLIE (CDD11) - }  \\ 

& \multicolumn{4}{c}{OKNet~\cite{cui2024omnioknet}}    & \multicolumn{4}{c}{\cite{weatherdiff}} & \multicolumn{4}{c}{OneRestore~\cite{guo2024onerestore}}  \\

\midrule

MUSIQ   & 67.832 & \underline{67.953} & 69.967 & 63.675 & 69.643 & 69.111 & 70.688 & 64.287 & 69.704 & \underline{70.050} & 71.607 & 65.578 \\

MANIQA  &  0.3809 & \underline{0.3965} & 0.4261 & 0.3328 & 0.3575 & \underline{0.3671} &  0.3941 & 0.3126 & 0.4277 & \underline{0.4531} & 0.4799 & 0.3619 \\

Q-Align & 3.9201 & \underline{4.1880} & 4.3936 & 3.6550 & 4.1281 & \underline{4.2604} & 4.4086 & 3.7904 & 4.1983 & \underline{4.3741} & 4.5656 & 3.9126 \\
NIQE $\downarrow$    & 2.7342 & 2.9758 & 2.9789 & 2.9956 & 2.7349 & 2.9613 & 2.9511 &  3.0477 & 3.3176& \textbf{3.3084} & 3.3547 & 3.2066 \\
CLIPIQA & 0.5032 &  \underline{0.5185} & 0.5786 & 0.4776 & 0.5184 & \underline{0.5322} &  0.5997 & 0.4816 & 0.6051 & 0.5862 & 0.6419 & 0.5409 \\
LIQE    & 3.4979 & \underline{4.1540} & 4.3748 & 3.4762 & 3.9674 & \underline{4.3408} &    4.4619 & 3.7231 & 4.0552 & \underline{4.4837} & 4.5790 & 3.8407 \\
\bottomrule
\end{tabular}%
\end{adjustbox}
\caption{
Quantitative comparisons across various IR models and IR tasks. We \underline{underline} the results where the optimized image $y_{\text{TTPO}}$ outperforms the restored image
$y_0$ in evaluation metrics. If $y_{\text{TTPO}}$ surpasses the preferred sample $y_w$, we further highlight the result in \textbf{bold}. Notably, we also provide numerical metric values for $y_w$ and $y_l$ in the Table as references.
}
\label{tab:main_result}
\end{table*}

\subsection{Optimization of Restored Images}
\label{sec:stage3}

The core of TTPO lies in optimizing the restored image.
In this stage, we leverage pre-trained diffusion models \cite{flux} and integrate reward signals from the selected preference samples \((y_w, y_l)\) via a conditioned diffusion process. This stage optimizes the initially restored image $y_0$, ultimately producing a perceptually enhanced version.

During the denoising process starting from noise, given the noisy data \(x_t\) at timestep \(t\), the model follows the pre-trained denoising trajectory (\ie, Eq.~\ref{equ:1}) to compute the update direction \(\Delta t \cdot v_\theta(x_t,t)\) for the next timestep. Conditional information can be intuitively incorporated into this trajectory by introducing an additional correction term to this update direction.
Note that the preference samples \(y_w, y_l\) are in the ``clean'' space at $t=0$.
Accordingly, we compute the conditional guidance \(\mathcal{L}_c(x_0,y_w,y_l)\) at each timestep by first deriving the predicted clean data \(x_0\), where \(x_0=x_t+t \cdot v_\theta(x_t,t)\), as defined in Eq.~\ref{equ:1}.
We incorporate preference information into the conditioning process by modulating the reward signals through the calculation of the distance between \(x_0\) and \(y_w, y_l\) as the relative rewards:
\begin{equation}
    \mathcal{L}_{\text{TTPO}}=-\log\sigma(\beta d(x_0, \mathcal{E}(y_w)) - \beta d(x_0, \mathcal{E}(y_l))),
    \label{eq:6}
\end{equation}
where \(d\) denotes a similarity-aware distance (\eg, cosine similarity as the reward), \(\beta\) controls the strength of the reward signals, and \(\mathcal{E}\) is VAE encoder. For distance metrics where lower values indicate greater similarity (\eg, L1), a negative sign is applied to \(d\) to ensure a consistent reward direction. Notably, we also include \(y_0\) itself as an additional condition to preserve structural consistency:
\begin{gather}
    \mathcal{L}_r= \| x_0-\mathcal{E}(y_0) \|^2_2,
    \label{eq:7}
    \\
    \mathcal{L}_c(x_0,y_0,y_w,y_l)=\alpha\mathcal{L}_{\text{TTPO}}+\mathcal{L}_r,
    \label{equ:8}
\end{gather}
where $\alpha$ is the weighting parameter. Next, we compute the correction term as:
\begin{equation}
    -\nabla_{x_t}\mathcal{L}_c=\frac{\partial \mathcal{L}_c}{\partial x_t}=\frac{\partial \mathcal{L}_c}{\partial x_0}\cdot\frac{\partial x_0}{\partial x_t}=\frac{\partial \mathcal{L}_c}{\partial x_0}.
\end{equation}
Here, we apply a stop gradient operation on \(v_\theta(x_t,t)\) for two reasons: (i) to introduce a correction term that guides the update of $x_t$ without interfering with the pre-trained denoising direction, and (ii) to avoid destabilizing the overall denoising process by altering the learned denoising trajectory. Detailed derivations are provided in Sec.~\ref{sec:algo}. Finally, we integrate the correction term into Eq.~\ref{equ:1} to compute the noisy data for the next timestep:
\begin{equation}
    x_{t'}=x_t+\Delta t\cdot v_\theta(x_t,t)-g\nabla_{x_t}\mathcal{L}_c,
\end{equation}
where $g$ is the scaling factor. Intuitively, integrating conditional guidance into the denoising process enables the restored image to better align with preferences by maximizing the reward for preferred samples and minimizing it for dispreferred ones, all while preserving structural consistency.

A potential issue arises when \(y_l\) closely resembles \(y_0\) and the structural consistency loss (Eq. \ref{eq:7}) causes $x_0$ to be similar to $y_l$, resulting in an artificially inflated reward in Eq.~\ref{eq:6}. This undermines the effectiveness of preference guidance and may result in unstable optimization.
To mitigate this, we apply frequency decomposition to \(x_0\) and separately compute preference optimization and structural preservation in distinct frequency components.
Specifically, we derive the preference optimization loss \(\mathcal{L}_{\text{TTPO}}\) from the high-frequency components, where textures are most prominent. Simultaneously, we enforce structural consistency $\mathcal{L}_r$ using the low-frequency components.
This process is formulated as:
\begin{multline}
    \mathcal{L}_{\text{TTPO}}= - \log \sigma \Bigl( \beta \mathcal{D} \bigl( \mathcal{F}^{\text{high}}(x_0),\mathcal{F}^{\text{high}}(\mathcal{E}(y_w)) \bigr) \\
    -\beta\mathcal{D} \bigl( \mathcal{F}^{\text{high}}(x_0),\mathcal{F}^{\text{high}}(\mathcal{E}(y_l)) \bigr) \Bigr),
\end{multline}
\begin{equation}
    \mathcal{L}_r=\|\mathcal{F}^{\text{low}}(x_0)-\mathcal{F}^{\text{low}}(\mathcal{E}(y_0))\|^2_2,
\end{equation}
where \(\mathcal{F}^{\text{high}}(\cdot)=\mathtt{FFT}(\cdot) \odot (1-\mathcal{G}(D_0))\) and \(\mathcal{F}^{\text{low}}(\cdot)=\mathtt{FFT}(\cdot) \odot \mathcal{G}(D_0)\). \(\mathcal{G}, D_0\) denote the Gaussian low-pass filter and the normalized cutoff frequency, respectively.

We divide the denoising process into three stages with distinct optimization objectives, inspired by findings that diffusion models generate structural information during early timesteps, while fine details are refined in later steps~\cite{rombach2022highStableDiffusion}. 
In the first stage $(T_1,1]$, only $\mathcal{L}_r$ is applied for structural guidance.
In the middle stage $(T_2, T_1]$, both $\mathcal{L}_{\text{TTPO}}$ and $\mathcal{L}_r$ are employed.
Finally, in the last stage $[0,T_2]$, only $\mathcal{L}_{\text{TTPO}}$ is used for fine-grained preference optimization.

\noindent\textbf{Discussion about Pixel Space v.s. Latent Space.} State-of-the-art diffusion models are commonly built upon Latent Diffusion Models~\cite{rombach2022highStableDiffusion}, which perform the denoising process in the latent space. While the pixel space can provide more precise spatial information, we opt to conduct conditioning in the latent space. This choice is motivated by two key factors: decoding to the pixel space at each timestep is computationally expensive, and, more importantly, we empirically observed that gradient backpropagation through the VAE decoder can be unstable.

\noindent\textbf{Comparing $\mathcal{L}_{\text{TTPO}}$ to $\mathcal{L}_{\text{DPO}}$.} While $\mathcal{L}_{\text{DPO}}$~\cite{rafailov2023directdpo} focuses on preference alignments during training, $\mathcal{L}_{\text{TTPO}}$ is designed for preference optimization at test time. Our method offers a more flexible and \textit{model-agnostic} solution by directly optimizing the restored image ($y_0$), eliminating the need to retrain IR models.

\begin{table*}[t]
\centering
\begin{adjustbox}{width=\textwidth}
\setlength{\tabcolsep}{10pt}
\begin{tabular}{@{}lcccccccccc@{}}
\toprule
        & $y_0$     & DDRM     & DDNM    & TAO   & $y_{\text{TTPO}}$      & $y_0$     & DDRM     & DDNM     & TAO    & $y_{\text{TTPO}}$      \\ 
        \cmidrule(lr){2-6}  \cmidrule(lr){7-11}
        & \multicolumn{5}{c}{(a) DN - Restormer~\cite{zamir2022restormer}}    & \multicolumn{5}{c}{(b) DN - DMID~\cite{dmiddenoising}}  \\ \midrule

MUSIQ   & 61.317 & 59.345 & \underline{60.509} & 58.930 & \textbf{61.054} & 61.696 & 59.304 & \underline{60.908} & 58.990 & \textbf{61.098} \\

MANIQA  & 0.4690 &  0.4450  &  \underline{0.4614}  &  0.4326  & \textbf{0.4741} & 0.4676 &  0.4437 & \underline{0.4611} & 0.4338 & \textbf{0.4767} \\

Q-Align &  2.8213 &  2.7105  & 2.7722  & \underline{2.7829} &  \textbf{2.9281} &   2.8686 & 2.7248 & 2.8037 & \underline{2.8245}  & \textbf{2.9687}\\
NIQE $\downarrow$    &  7.0202 &  7.7514  &  7.6399 & \textbf{6.6802} &  \underline{7.0108} &  6.7982     &  7.7266 & 7.5267 & \textbf{6.6481} & \underline{6.8480}   \\
CLIPIQA & 0.5600 &  0.5153 & 0.5225 & \textbf{0.6064} & \underline{0.5849} &  0.5802 & 0.5365 & 0.5420 & \textbf{0.6129} & \underline{0.6011} \\
LIQE    &  3.8615 &  3.6078  &  3.6512 & \underline{4.1011}   & \textbf{4.2702} & 3.9228      & 3.6481 & 3.6975 & \underline{4.1588} & \textbf{4.3020}     \\
\bottomrule
\end{tabular}%
\end{adjustbox}
\caption{Quantitative comparisons between TTPO and previous ZSDIR methods, DDRM~\cite{kawar2022denoisingddrm}, DDNM~\cite{wang2023zeroddnm}, and TAO~\cite{gou2024testTAO}. Results are reported on Kodak24~\cite{kodak} with the $256\times 256$ resolution. For a fair comparison, we use the restored image $y_0$ (rather than the distorted image) as input to all ZSDIR methods. Each method is executed multiple times to identify the optimal hyperparameter settings. The best and second-best results are \textbf{bolded} and \underline{underlined} (values of $y_0$ serve as reference).}
\label{tab:zsdir_result}
\end{table*}

\section{Experiments}

\subsection{Parameter Settings
\label{sec_parameter_settings}}
\textbf{Candidate Preference Images Generation.} To increase the diversity of generated preference candidates, we employ SD2.1~\cite{rombach2022highStableDiffusion}, SD3~\cite{esser2024scalingsd3}, and FLUX~\cite{flux} as pre-trained diffusion models. We set $s$ to [0.1, 0.15, 0.2, 0.25, 0.3] to generate five samples for each diffusion model. There are 16 candidates for each restored image (\ie, $N=16$) in total, including the restored image itself. \textbf{Preference Image Selection.} We adopt the combination of MUSIQ~\cite{ke2021musiq}, MANIQA~\cite{yang2022maniqa}, and Q-Align~\cite{wu2023q} as substituted IQA metrics.  \textbf{Preference Optimization.} Unless stated otherwise, all diffusion models used for conditional denoising in this paper are FLUX~\cite{flux}, due to its superior generative performance. We empirically set the hyperparameters $\alpha, \beta, g$ as 0.5, 1, and 1000, respectively. We use MSE loss for $\mathcal{L}_r$ and set the distance metric in $\mathcal{L}_{\text{TTPO}}$ as L1 loss. For the cutoff frequency $D_0$, we set it to 0.9. Referring to previous work~\cite{gou2024testTAO}, we divide the denoising process into three stages by setting $T_1=0.7$ and $T_2=0.1$. All experiments are conducted on two GPUs with 24G VRAM and 35.58 TFLOPS. More details for diffusion models are provided in Sec.~\ref{sec:parameter}.

\subsection{Results on Various IR Models and IR Tasks}
\label{sec:results}

We consider several representative IR tasks to demonstrate the effectiveness of TTPO: (i) the most classical IR task, image denoising (DN); (ii) the classical image super-resolution (CISR) task; 
(iii) the practical real-world image super-resolution (RISR) task;
(iv) the image deraining (DR) task; (v) low-light image enhancement (LLIE) task; and
(vi) the zero-shot diffusion-based IR (ZSDIR) task. \textit{Since our goal is to optimize the restored image, we primarily focus on methods whose visual results are publicly available.}

\begin{figure}
    \centering
    \includegraphics[width=1\linewidth]{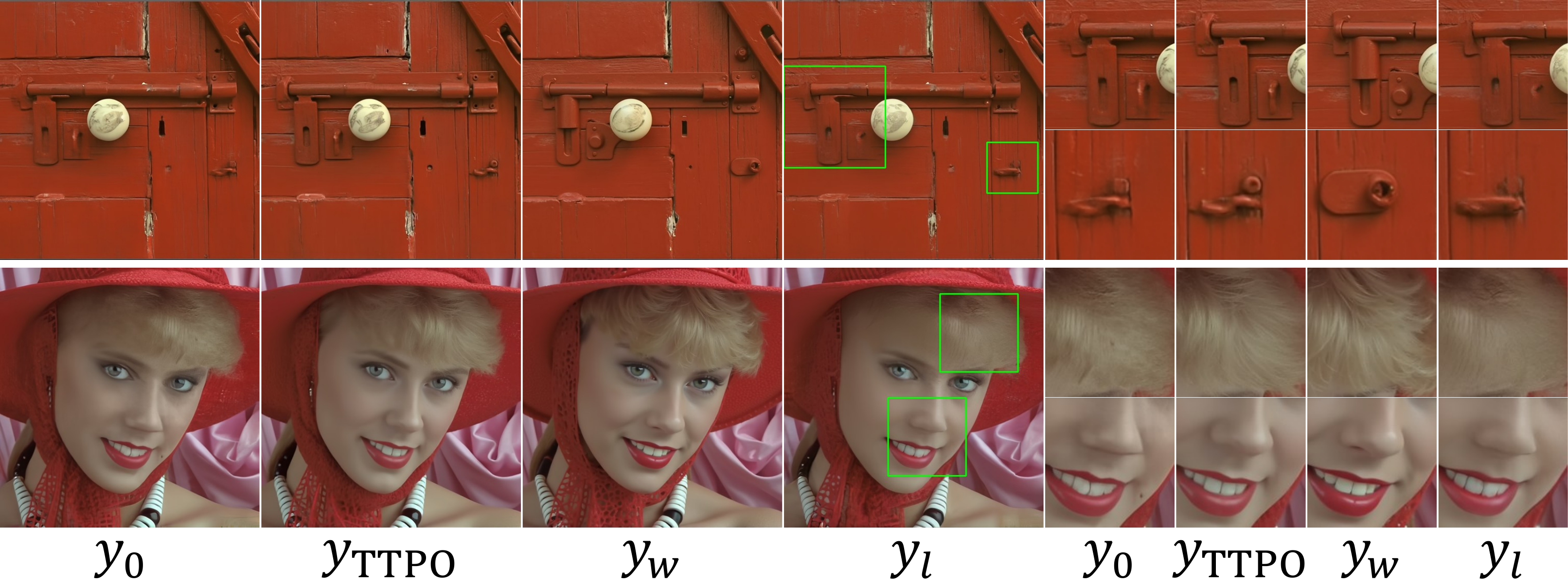}
    \caption{Qualitative comparisons between the initially restored image $y_0$, optimized image $y_{\text{TTPO}}$, preferred image $y_w$, and dispreferred image $y_l$.}
    \label{fig:mainexp}
\end{figure}

\noindent \textbf{Main Results.} We present the quantitative results in Tab.~\ref{tab:main_result}. We report MUSIQ~\cite{ke2021musiq}, MANIQA~\cite{yang2022maniqa}, and Q-Align~\cite{wu2023q} in the Table, where these metrics are leveraged to select the preference images. Additionally, we include three more commonly adopted NR-IQA metrics, NIQE~\cite{niqe}, CLIPIQA~\cite{clipiqa}, and LIQE~\cite{liqe}, to provide a comprehensive evaluation of the optimized images. Several conclusions can be drawn from the Table: (i) TTPO consistently improves over the restored image $y_0$ 
across almost all tasks and evaluation metrics. (ii) In many cases, TTPO achieves improvements not only on the metrics used during the preference selection stage but also on additional metrics that were not involved in the optimization. (iii) In some tasks, TTPO even surpasses the preferred image $y_w$. These three observations demonstrate the flexibility and effectiveness of TTPO against various IR models/tasks. We demonstrate the qualitative comparisons in Fig.~\ref{fig:mainexp}. As observed, images optimized through preference guidance exhibit more realistic textures. For example, $y_{\text{TTPO}}$ presents more natural representations of hair and teeth, without artifacts such as twisted lines. Notably, although $y_w$ sometimes demonstrates finer texture details, it often introduces undesirable structural changes (\eg, structure shifts in the door region). Therefore, $y_w$ \textit{cannot be directly used as the final result}, as structural consistency is a fundamental requirement in IR. In contrast, $y_{\text{TTPO}}$ retains structural information with improved perceptual quality. 

\begin{figure}[t]
    \centering
    \begin{subfigure}[b]{0.46\textwidth}
        \centering
        \includegraphics[width=\textwidth]{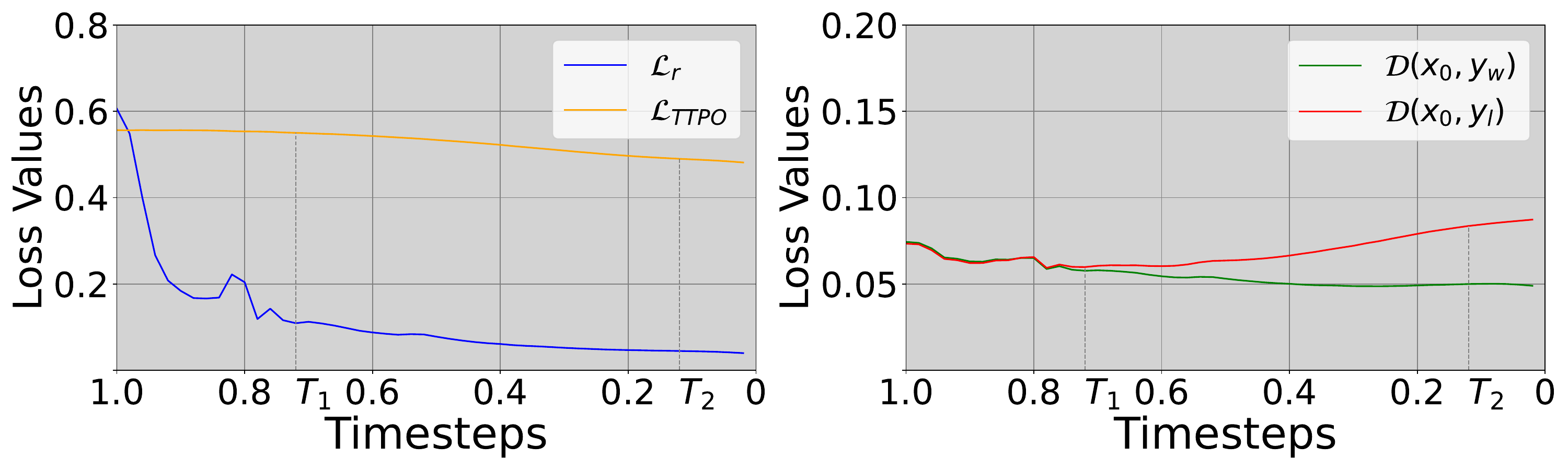} 
        \label{fig:subfig1}
    \end{subfigure}
    \hspace{0.06\textwidth} 
    \begin{subfigure}[b]{0.46\textwidth}
        \centering
        \includegraphics[width=\textwidth]{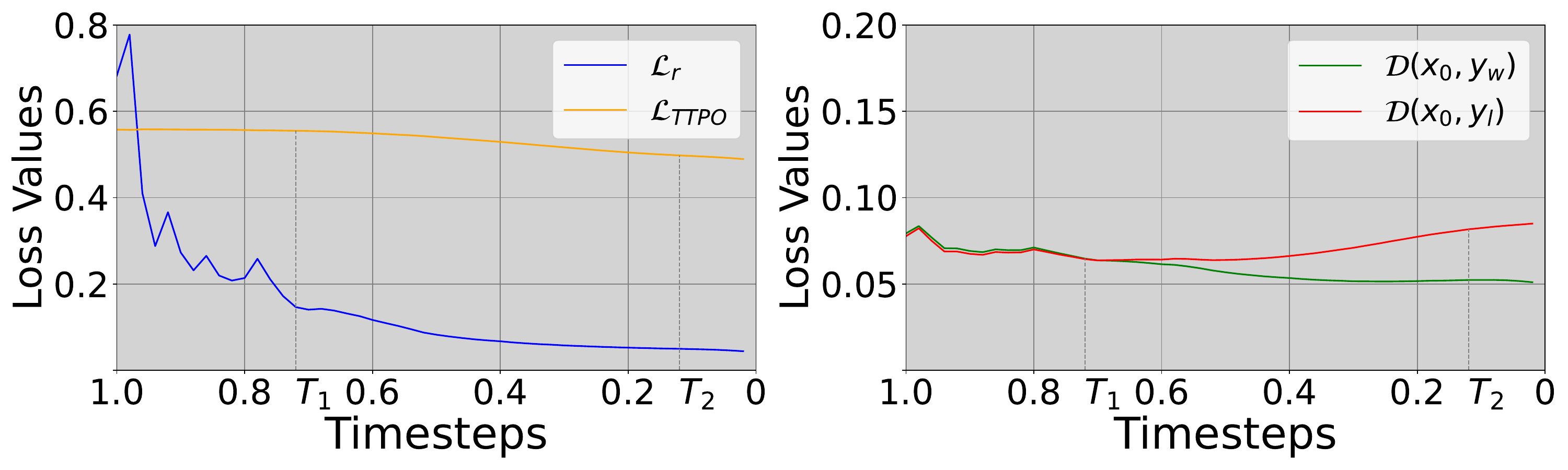} 
        \label{fig:subfig2}
    \end{subfigure}
    \caption{Loss curves for two images in Fig.~\ref{fig:mainexp} across all diffusion timesteps.}
    \label{fig:curves}
\end{figure}

\noindent \textbf{Comparing TTPO with ZSDIR Methods.} Since ZSDIR methods are based on pre-trained diffusion models and do not require training, we compare our TTPO with them for \textit{optimizing} restored images. We include three representative ZSDIR baselines: DDRM~\cite{kawar2022denoisingddrm}, DDNM~\cite{wang2023zeroddnm}, and TAO~\cite{gou2024testTAO}. Specifically, the restored image $y_0$ is used as input for all methods. For a fair comparison, all inputs are resized to $256 \times 256$, including those for TTPO. Each ZSDIR method is executed multiple times to determine the optimal hyperparameter settings. As shown in Tab.~\ref{tab:zsdir_result}, TTPO achieves the best or second-best results across all metrics, which indicates the effectiveness of preference guidance.

\begin{figure}
    \centering
    \includegraphics[width=\linewidth]{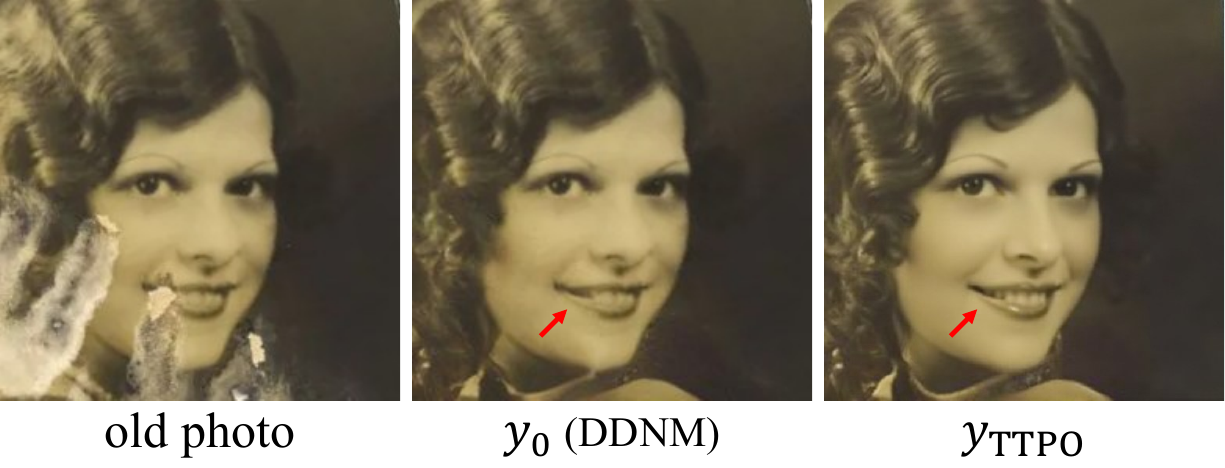}
    \caption{Integrating TTPO with the existing ZSDIR method DDNM~\cite{wang2023zeroddnm}.}
    \label{fig:oldphoto}
\end{figure}

\noindent \textbf{Integrating TTPO with ZSDIR Methods.} Notably, ZSDIR methods can be seamlessly integrated with TTPO to form a two-stage test-time restoration and optimization method for distorted images. We provide an example in Fig.~\ref{fig:oldphoto}, where the old photo is first restored by DDNM~\cite{wang2023zeroddnm} and optimized by TTPO. As shown, TTPO enhances the texture details and brings the old photo back to life.

\begin{figure}
    \centering
    \includegraphics[width=1\linewidth]{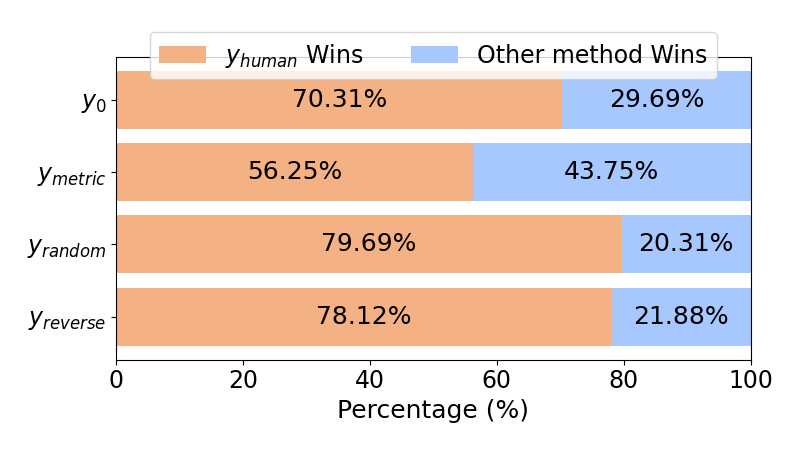}
    \caption{User study of $y_{\text{human}}$ \textit{vs.} other selection methods.}
    \label{fig:human}
\end{figure}

\noindent \textbf{User Study.} To further demonstrate the preference optimization capability of TTPO, we conduct a user study on Kodak24~\cite{kodak} dataset. We invite eight human experts to participate in this study, each with more than three years of research experience in low-level vision. Each expert is first asked to select preference images in stage two. We then optimize restored images based on their selections. Subsequently, these experts participate in stage three to compare the optimized images $y_{\text{human}}$ with the following settings: (i) $y_0$; (ii) $y_{\text{metric}}$ optimized by metric-selected preference images; (iii) $y_{\text{random}}$ optimized by randomly selected preference images; and (iv) $y_{\text{reverse}}$ optimized by switching the preference image pairs. We report the win rate of $y_{\text{human}}$ in Fig.~\ref{fig:human}. As observed, $y_{\text{human}}$ surpasses $y_0$ in most comparisons, which indicates the ability of TTPO to enhance the subjective quality of $y_0$. Additionally, a win rate near 50\% between $y_{\text{human}}$ and $y_{\text{metric}}$ suggests that the NR-IQA metrics employed in TTPO align well with human preferences. More settings are detailed in Sec.~\ref{sec:userstudy}.

\noindent \textbf{Analysis of Conditions.} We illustrate the curves of $\mathcal{L}_{\text{TTPO}}, \mathcal{L}_r, \mathcal{D}(x_0,\mathcal{E}(y_w)),$ and $\mathcal{D}(x_0,\mathcal{E}(y_l))$ in Fig.~\ref{fig:curves}. As demonstrated, both the preference guidance loss $\mathcal{L}_{\text{TTPO}}$ and the structural consistency loss $\mathcal{L}_r$ gradually decrease throughout the denoising process. Meanwhile, the distance between the clean data and the preferred image becomes smaller, while the distance to the dispreferred image grows larger. This demonstrates the effectiveness and correctness of the TTPO pipeline.

\begin{figure}
    \centering
    \includegraphics[width=\linewidth]{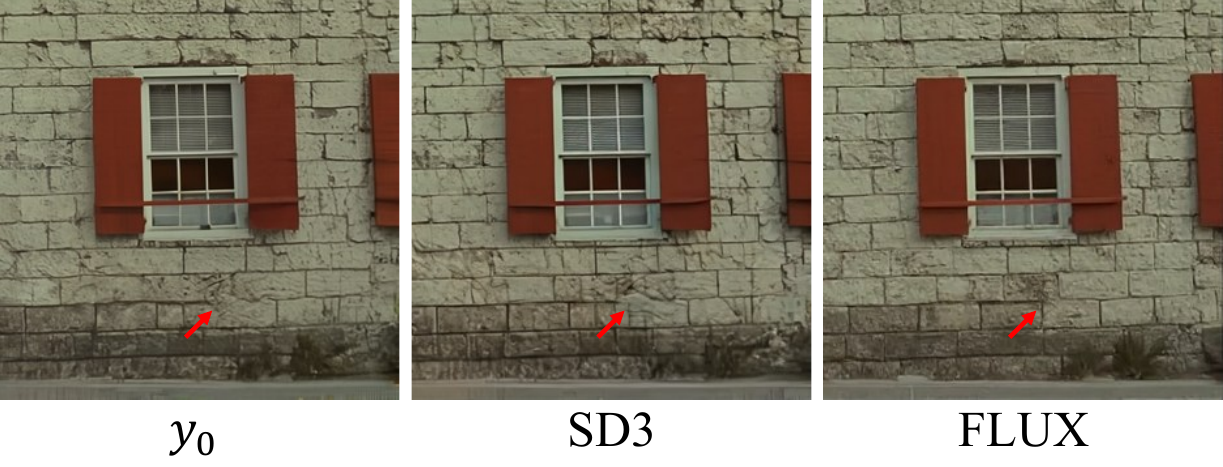}
    \caption{Qualitative comparisons between SD3 and FLUX utilized in the third stage.}
    \label{fig:abmodel}
\end{figure}

\subsection{Ablation Studies}
\label{sec:ablationstudy}
Unless stated otherwise, all experiments in this part are conducted on Kodak~\cite{kodak} denoised (DN) by SwinIR~\cite{liang2021swinir}. We provide more ablation studies about parameter settings in Sec.~\ref{sec:moreablation}.

\noindent \textbf{Effectiveness of $\mathcal{L}_{\text{TTPO}}$, Frequency Decomposition, and Division of Denoising Steps.} As demonstrated in Tab.~\ref{tab:abloss}, $\mathcal{L}_{\text{TTPO}}$ brings obvious improvements against (i). This indicates that our developed preference guidance effectively transfers the preference as a reward signal to guide the denoising process. TTPO achieves better performance with frequency decomposition in (iii), as it prevents the reward for $y_l$ from being excessively high due to the similarity in low-frequency structural information. By dividing the denoising process into three stages and applying different guidance in (iv), the performance increases compared to (ii), as expected. This is because introducing preference guidance too early may interfere with the generation of structural details, leading to a suboptimal denoising process. By putting things together in (v), TTPO achieves the best performance, indicating the effectiveness of our optimization pipeline.

\noindent \textbf{Flexibility of Diffusion Model in the Third Stage.} Since the TTPO pipeline does not require any retraining, it provides flexibility in choosing the diffusion model used to align restored images with human preferences. We compare the performance between FLUX~\cite{flux} and SD3~\cite{esser2024scalingsd3} in Tab.~\ref{tab:abmodel} and Fig.~\ref{fig:abmodel}. As demonstrated, although SD3 achieves slightly better results on IQA metrics, it suffers from unnatural texture generation (\eg, twisted lines). However, since FLUX has more parameters than SD3, the latter can perform the TTPO process more efficiently, resulting in faster optimization. Therefore, users can choose the most appropriate diffusion model for preference optimization, depending on their trade-off between speed and image quality.

\begin{table}[H]
    \centering
    \small
    \begin{adjustbox}{width=\hsize}
    \setlength{\tabcolsep}{10pt}
    \begin{tabular}{@{}lcccc@{}}
    \toprule
    Model & MUSIQ & MANIQA & Q-Align & Inf. Time \\ \midrule
    SD3   &  74.709 &  0.4910 & 4.1135  &  26s\\
    FLUX  & 74.242   &   0.4912    &  4.3201 &  45s  \\ \bottomrule
    \end{tabular}%
    \end{adjustbox}
    \caption{Different diffusion models used in the third stage.}
    \label{tab:abmodel}
\end{table}

\begin{table}[H]
    \centering
    \begin{adjustbox}{width=\hsize}
    \setlength{\tabcolsep}{4pt}
    \begin{tabular}{@{}cccccccc@{}}
    \toprule
    & $\mathcal{L}_r$ & $\mathcal{L}_{\text{TTPO}}$ & \texttt{FFT} & $T_1,T_2$ & MUSIQ & MANIQA & Q-Align \\ \midrule
     (i) & \checkmark  &   & & & 73.437 & 0.4701  & 4.1768     \\
      (ii) & \checkmark &   \checkmark    &  & &  73.895 &   0.4798 &  4.2397      \\
     (iii) & \checkmark &   \checkmark    & \checkmark & &   74.103 &  0.4857 &  4.2422     \\
    (iv) &  \checkmark &   \checkmark    &  & \checkmark &  74.051 & 0.4835 & 4.2585     \\
     (v) & \checkmark &   \checkmark &   \checkmark &   \checkmark   &    \textbf{74.242}   &   \textbf{0.4912}     &  \textbf{4.3201}       \\
      \bottomrule
    \end{tabular}
    \end{adjustbox}
    \caption{Effectiveness of each component in the third stage.}
    \label{tab:abloss}
\end{table}

\section{Conclusion\label{sec_conclusion}}

In this paper, we propose the first Test-Time Preference Optimization (TTPO) paradigm for image restoration. By generating preference data on-the-fly and leveraging a three-stage optimization pipeline, TTPO effectively aligns restored images with user preferences, eliminating the need for costly model retraining or labor-intensive data collection. Extensive experiments highlight the flexibility and effectiveness of TTPO across diverse image restoration tasks and backbones, demonstrating its capability to optimize images in various scenarios while enhancing perceptual quality and preserving structural consistency.
Our method, however, has two main limitations: (i) the inference process of FLUX requires considerable computational resources, leading to relatively long runtime; and (ii) current NR-IQA metrics still fall short of fully capturing human preferences, making it difficult for metric-based selection to fully align the optimization results with human judgments. We leave improving the computational efficiency of TTPO and developing more accurate preference selection mechanisms as future works.

\section*{Acknowledgments}
This work was supported in part by NSFC under Grant 623B2098, 62021001, and 62371434, the Postdoctoral Fellowship Program of CPSF under Grant Number GZC20252293, and the China Postdoctoral Science Foundation-Anhui Joint Support Program under Grant Number 2024T017AH. This work was also funded by Anhui Postdoctoral Scientific Research Program Foundation (No.2025A1015) and the Fundamental Research Funds for the Central Universities under Grant Number WK2100250064.

\appendix

\section*{Appendix}

In the appendix, we first explain the details for NR-IQA metrics selection in Sec.~\ref{sec:nriqa}. An algorithm of TTPO is demonstrated in Sec.~\ref{sec:algo}. More details of the user study and parameter settings for diffusion models utilized in this paper are provided in Sec.~\ref{sec:userstudy} and Sec.~\ref{sec:parameter}, respectively. As a complement to the experimental results presented in the main paper, we list additional quantitative and qualitative results in Sec.~\ref{sec:moreresults} and Sec.~\ref{sec:moreablation}. Finally, we discuss limitations of this work in Sec.~\ref{sec:limi}.

\section{NR-IQA Metrics Selection with Subjective Experiments}
\label{sec:nriqa}

The ideal approach for selecting preference images would involve direct human evaluation, as adopted in the user study (Sec.~\ref{sec:ablationstudy}). However, this process becomes labor-intensive and impractical when the number of candidate images increases significantly. To address this, we propose a simple yet effective alternative by leveraging no-reference image quality assessment (NR-IQA) metrics to rank candidate images. Recent NR-IQA methods~\cite{yang2022maniqa,clipiqa,liqe,wu2023q} have demonstrated commendable assessment capabilities without requiring reference images, making them particularly suitable for our test-time scenario where ground-truth data is typically unavailable. Nevertheless, these NR-IQA metrics still exhibit noticeable misalignment with human preferences~\cite{zhang2022perceptualnriqa,bsrgan}. To address this, we propose to select top-aligned metrics with subjective experiments and utilize the combination of selected NR-IQA metrics~\cite{chen2024restoreagent,lin2025jarvisir,peng2021multi} to improve the alignments.

\begin{table}[h]
\centering

\resizebox{1\linewidth}{!}{%
\begin{tabular}{@{}ccccccl@{}}
\toprule
        & MUSIQ & NIQE & MANIQA & CLIPIQA & LIQE & Q-Align \\ \midrule
``matches'' & 338/500 & 279/500 & 345/500 & 108/500 &    286/500  &  366/500    \\ \bottomrule
\end{tabular}%
}
\caption{Subjective experiments for NR-IQA metrics selection.}
\label{tab:subjectexp}
\end{table}

Specifically, we randomly sample 50 images from the restored outputs of various IR models, each processed using the same candidate image generation method described in Sec.~\ref{sec:stage1}. For each group of candidate images, we manually select the preferred and dispreferred samples, which are then used as ground-truth labels for preference evaluation. We consider six popular NR-IQA metrics from \texttt{pyiqa} Python package, including MUSIQ~\cite{ke2021musiq}, NIQE~\cite{niqe}, CLIPIQA~\cite{clipiqa}, MANIQA~\cite{yang2022maniqa}, LIQE~\cite{liqe}, and Q-Align~\cite{wu2023q}. To balance evaluation reliability and efficiency, we employ a combination of three NR-IQA metrics for assessing the quality of each image. For each group of images, if the preferred/dispreferred samples selected by the metric-based ranking match the human-annotated ground-truth preferences, and a given NR-IQA metric is part of the selected combination, we record a ``match'' for that metric. This allows us to measure how frequently each metric contributes to preference-aligned decisions and select the top metrics as substitutes for humans. Generally, each group of candidate preference images will be assessed by $\text{C}_6^3=20$ combinations of metrics. For each metric, it will be included in 10 combinations (\ie, $\text{C}_{6-1}^{3-1} = 10$). We report the ``matches'' for each NR-IQA metric across 50 images in Tab.~\ref{tab:subjectexp}. As demonstrated, Q-Align, MANIQA, and MUSIQ are selected as the top three metrics most aligned with human preferences, and are leveraged as substitutes for human judgment in the preference image selection stage.

\begin{algorithm*}
\caption{Test-Time Preference Optimization (TTPO)}
\label{alg:ttpo}
\begin{algorithmic}[1]
\State \textbf{Input:} Restored image $y_0$, pre-trained diffusion model with VAE encoder $\mathcal{E}$ and VAE decoder $\mathcal{D}$, noise levels $\{s_k\}_{k=1}^N$
\State \textbf{Output:} Preference optimized image $\hat{y}_0$
\vspace{0.5em}
\State \# \textbf{Stage 1: Preference Image Generation}
\State Encode $x_0 \gets \mathcal{E}(y_0)$
\For{$k = 1$ to $N$}
    \State Add noise: $x_s^k \gets (1 - s_k)x_0 + s_k \epsilon$ \Comment{Eq. (2)}
    \State Generate edited image $y_0^k \gets \text{Denoise}(x_s^k)$ \Comment{Eq. (1)}
\EndFor
\State Form candidate preference images set $D = \{y_0, y_0^1, ..., y_0^N\}$
\vspace{0.5em}

\State \#  \textbf{Stage 2: Preference Image Selection}
\For{each $y_0^k \in D$}
    \State Compute hybrid normized NR-IQA score $r_k = \frac{\text{IQA}_1 + \text{IQA}_2 + \text{IQA}_3}{3}$ \Comment{Eq. (5)}
\EndFor
\State $y_w \gets \text{argmax}_k\ r_k$, \quad $y_l \gets \text{argmin}_k\ r_k$
\vspace{0.5em}

\State \# \textbf{Stage 3: Restored Image Optimization}
\For{each denoising timestep $t$}
    \State Predict $\hat{x}_0 = x_t + t \cdot v_\theta(x_t, t)$
    \State Compute $\mathcal{L}_{\text{TTPO}}$ using high-frequency components \Comment{Eq. (11)}
    \State Compute $\mathcal{L}_r$ using low-frequency components \Comment{Eq. (12)}
    \State Compute $\mathcal{L}_c = \alpha \mathcal{L}_{\text{TTPO}} + \mathcal{L}_r$
    \State Compute correction term: $-\nabla_{x_t}\mathcal{L}_c=\frac{\partial \mathcal{L}_c}{\partial \hat{x}_0}\cdot\frac{\partial \hat{x}_0}{\partial x_t}=\frac{\partial \mathcal{L}_c}{\partial \hat{x}_0}\cdot \frac{\partial (x_t+t\cdot\texttt{SG}(v_\theta(x_t, t)))}{\partial x_t} = \frac{\partial \mathcal{L}_c}{\partial \hat{x}_0}$
    \State Update latent: $x_{t'} \gets x_t + \Delta t \cdot v_\theta(x_t, t) - g \cdot \nabla_{x_t} \mathcal{L}_c$ \Comment{Eq. (10)}
\EndFor
\State Decode final output $\hat{y}_0 \gets \mathcal{D}(x_0)$
\end{algorithmic}
\end{algorithm*}

\section{Algorithm of TTPO}
\label{sec:algo}

The complete algorithm for TTPO is provided in Alg.~\ref{alg:ttpo}.

\noindent\textbf{Derivation of Eq. (9).} Notably, we apply a Stop Gradient (\texttt{SG}) operation on \(v_\theta(x_t,t)\) for two reasons: (i) we aim to introduce a correction term to guide the update of $x_t$, without interfering with the pre-trained denoising direction; and (ii) modifying the learned denoising trajectory may destabilize the overall denoising process. Hence, $\frac{\partial \hat{x}_0}{\partial x_t}$ is identically equal to 1, since $\texttt{SG}(v_\theta(x_t, t))$ is detached from the computation graph and does not contribute any gradient.

\noindent\textbf{Normalization of IQA Scores.} As the value ranges of different IQA metrics vary significantly, directly summing their scores can lead to an imbalance in their contributions, where some metrics may disproportionately influence the final reward. To address this, we follow~\cite{chen2024restoreagent,lin2025jarvisir} to apply Z-score normalization on each metric before summation: $\mathcal{S} = \frac{s - \mu}{\sigma}$, where $s$ denotes the original score, $\mu$ and $\sigma$ are the mean and standard deviation of $s$ across the candidate preference images set, respectively.

\section{More Details about User Study}
\label{sec:userstudy}

\begin{figure*}
    \centering
    \includegraphics[width=1\linewidth]{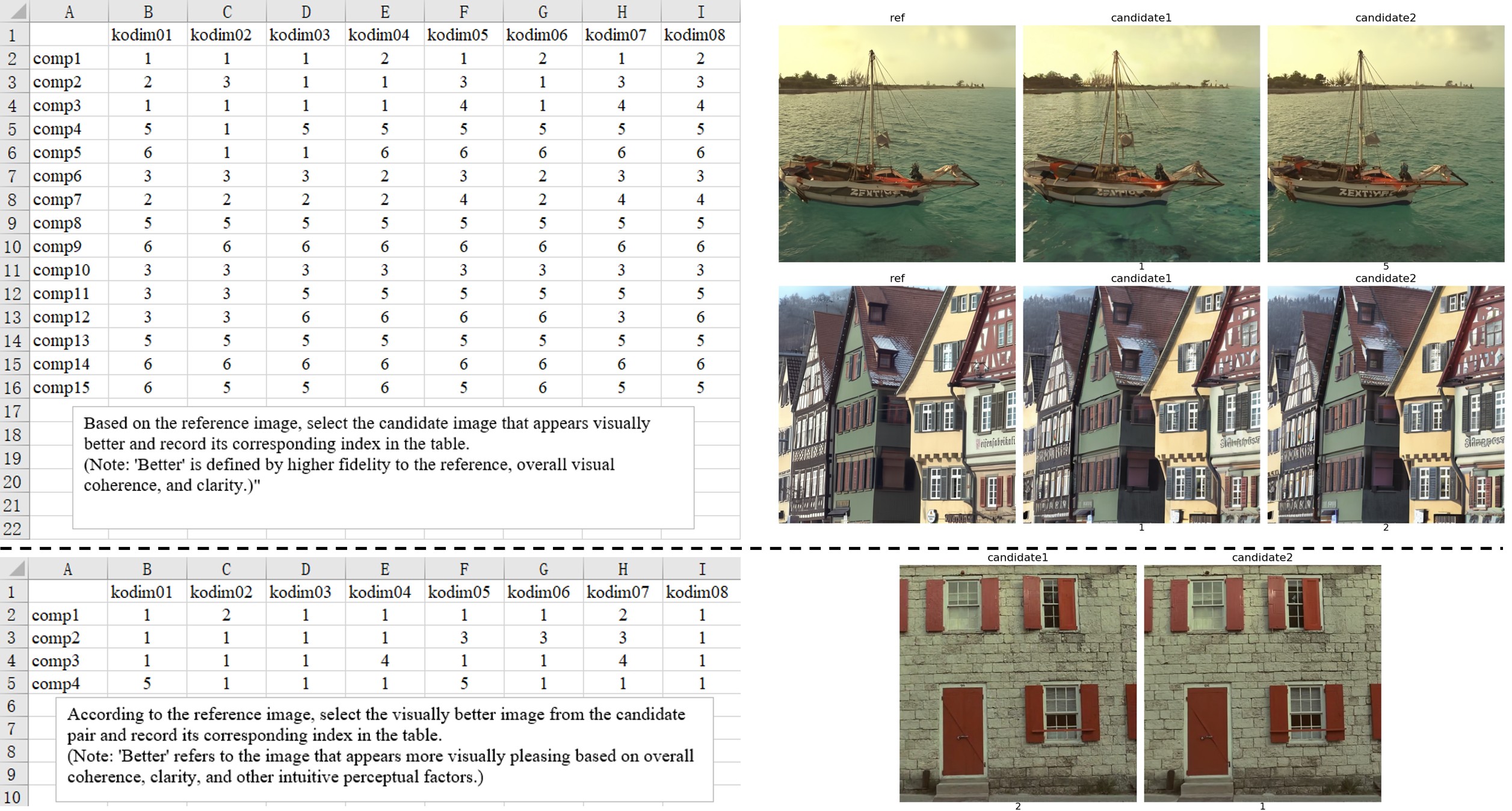}
    \caption{Screenshots of the user study: round one (top left) and round two (bottom left). Examples of visual comparisons (right).}
    \label{fig:userstudy}
\end{figure*}

In Sec. 5.2 of the main paper, we conduct a user study to demonstrate the capability of TTPO to optimize the restored image according to human preference. To simplify the selection process, we only adopt [0.15, 0.25] as $s$ for the generation stage. With three different diffusion models, \ie, SD2.1, SD3, and FLUX, there are $2\times3=6$ candidate preference images for each restored image. Eight proficient low-level experts are participating in this study. Each expert is asked to perform in stages two and three of TTPO. Specifically, the experts choose the better image between each pairwise comparison among all candidate preference images ($\text{C}_6^2 = 15$ groups in total) in the second stage of TTPO. Based on the number of selections, the most/least frequently chosen image (out of $8\times 15=120$ selections) is designated as the preferred/dispreferred image. Subsequently, we use the expert-annotated preference pairs to perform the third stage of TTPO. In the second round of the study, each expert is asked to compare the image optimized using human-annotated preferences with the outputs from the other four settings. For each pairwise comparison, if the human preference-optimized image is favored by the expert over the alternative, we record it as a win for $y_{\text{human}}$. Finally, the win rate of $y_{\text{human}}$ is computed as the ratio of the number of wins to the total number of pairwise comparisons, which in our setting is $8\times8=64$. We provide screenshots of these two rounds in Fig.~\ref{fig:userstudy}. 

\section{Details of Parameter Settings for Diffusion Models}
\label{sec:parameter}

\noindent\textbf{Generation Stage.} We provide the parameter settings for candidate preference images generation:
\begin{itemize}
    \item \textbf{Number of denoising steps:} We uniformly set the total denoising steps to 50 for SD2.1, SD3, and FLUX.
    \item \textbf{Height and width:} Since recent diffusion models are typically trained on images with resolutions larger than $512 \times 512$, we adopt a dynamic padding strategy to pad restored images to a spatial resolution greater than $512 \times 512$. Specifically, for images with resolutions smaller than $1024 \times 1024$, we pad them to the closest resolution among the set \(\{512 \times 512, 512 \times 768, 768 \times 512, 768 \times 768, 768 \times 1024, 1024 \times 768, 1024 \times 1024\}\). For images with resolutions larger than $1024 \times 1024$, we pad the images so that both height and width are multiples of 16, which helps to reduce GPU memory consumption.
    \item \textbf{Strength:} As stated in the main paper, we adopt a list of noise scales $s$ among [0.1, 0.15, 0.2, 0.25, 0.3] for each diffusion model.
    \item \textbf{Prompt:} We apply a universal prompt: ``Cinematic, high-contrast, photo-realistic, 8k, ultra HD, meticulous detailing, hyper sharpness, perfect without deformations,'' which is adopted from InvSR~\cite{yue2024arbitraryinvsr}.
    \item \textbf{Guidance Scale:} We use the default editing guidance scale for each diffusion model, \ie, 7.5 for SD2.1; 7.0 for SD3 and FLUX.
    \item \textbf{Seed:} A fixed random seed of 666666 is used consistently across all generation processes to ensure reproducibility.
\end{itemize}
\noindent \textbf{Optimization Stage.} We adopt the same settings as in the generation stage, except that for FLUX, the default denoising guidance scale is set to 3.5 and the strength parameter is not required.

\section{More Results of TTPO across Different IR Datasets}
\label{sec:moreresults}

\begin{figure*}
    \centering
    \includegraphics[width=0.9\linewidth]{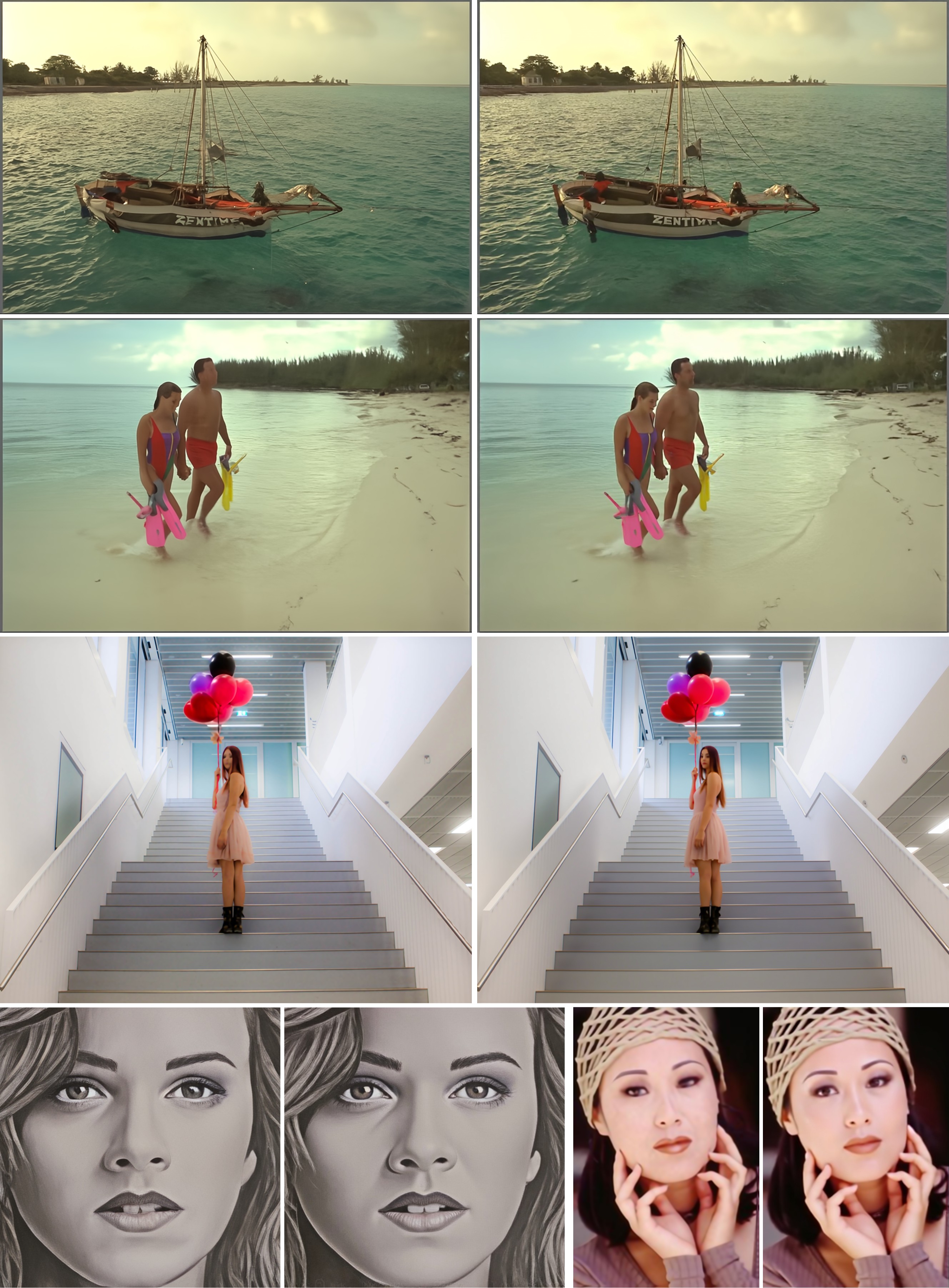}
    \caption{More qualitative comparisons on images with larger resolutions. Our TTPO theoretically supports preference optimization at any resolution. Left/Right is the restored/optimized image, respectively. Zoom in for better views.}
    \label{fig:supp2}
\end{figure*}

\begin{figure*}
    \centering
    \includegraphics[width=0.82\linewidth]{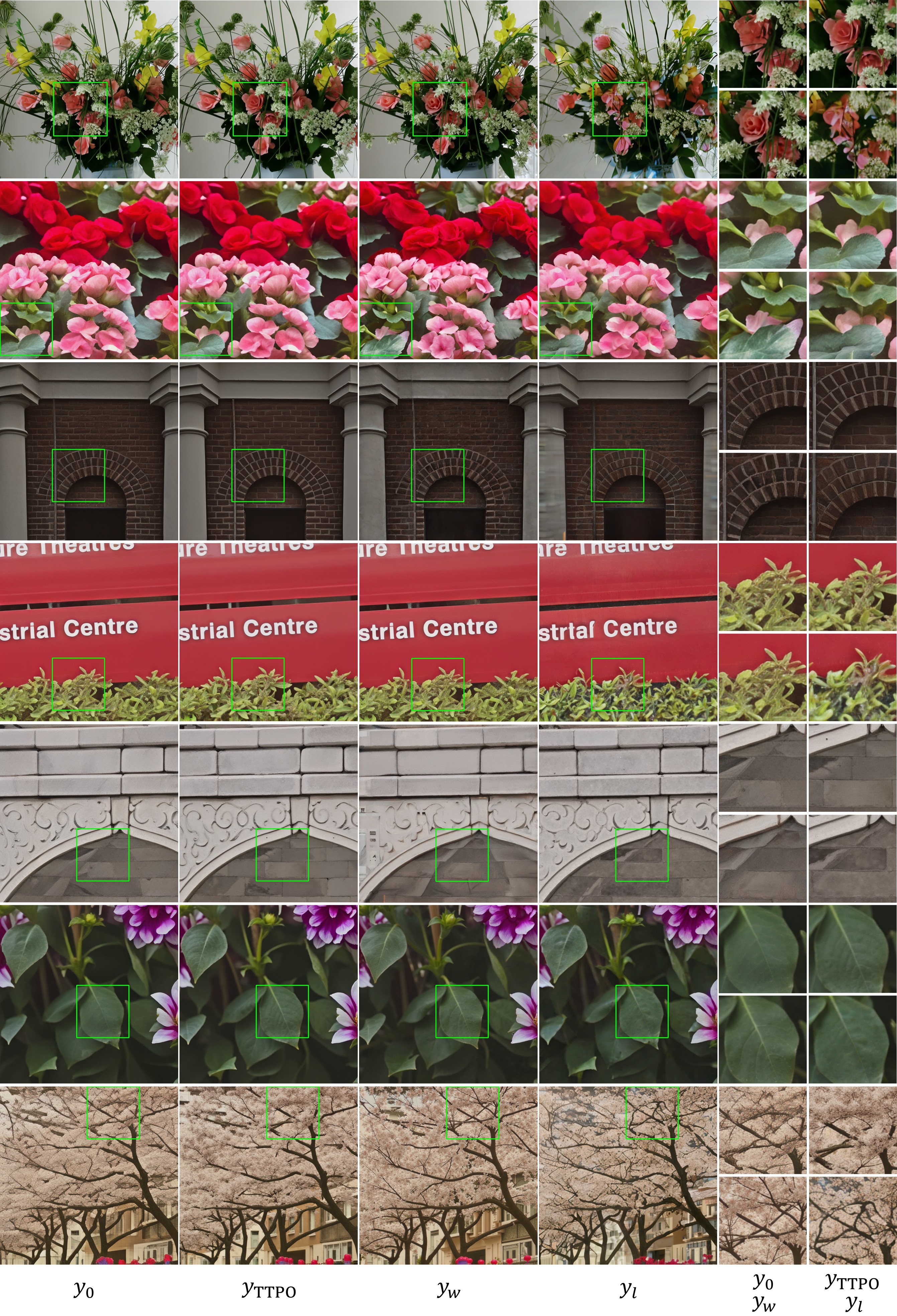}
    \caption{More qualitative comparisons on various IR datasets including Kodak~\cite{kodak}, McMaster~\cite{mcmaster}, RealSR~\cite{RealSR}, and DRealSR~\cite{wei2020componentdrealsr}.}
    \label{fig:supp}
\end{figure*}

To provide a more comprehensive overview beyond Tab. 1 of the main paper, we additionally present TTPO’s performance across different IR datasets in Tab.~\ref{tab:additional}, including McMaster~\cite{mcmaster}, CBSD68~\cite{cbsd}, Urban100~\cite{urban100}, Set5~\cite{Set5}, DRealSR~\cite{wei2020componentdrealsr}, nad DIV2K-Val~\cite{DIV2K}. As demonstrated, TTPO remains effective across various IR datasets, further substantiating its capability for preference optimization towards any IR models. We provide more qualitative results in Fig.~\ref{fig:supp2} and Fig.~\ref{fig:supp}. As illustrated in the figures, images optimized by TTPO exhibit textures that are more aligned with human preference. In some cases, TTPO even corrects artifacts introduced by super-resolution and enhances facial details.

\begin{table*}[]
\centering
\resizebox{\textwidth}{!}{%
\begin{tabular}{@{}lcccccccccccc@{}}
\toprule
        & \makebox[0.08\textwidth][c]{$y_0$}     & \makebox[0.08\textwidth][c]{$y_{\text{TTPO}}$}     & \makebox[0.08\textwidth][c]{$y_w$}     & \makebox[0.08\textwidth][c]{$y_l$}     & \makebox[0.08\textwidth][c]{$y_0$}     & \makebox[0.08\textwidth][c]{$y_{\text{TTPO}}$}     & \makebox[0.08\textwidth][c]{$y_w$}     & \makebox[0.08\textwidth][c]{$y_l$}     & \makebox[0.08\textwidth][c]{$y_0$}     & \makebox[0.08\textwidth][c]{$y_{\text{TTPO}}$}     & \makebox[0.08\textwidth][c]{$y_w$}     & \makebox[0.08\textwidth][c]{$y_l$}     \\ 
        \cmidrule(lr){2-5}  \cmidrule(lr){6-9} \cmidrule(lr){10-13}
        & \multicolumn{4}{c}{(a) DN, SwinIR, McMaster}    & \multicolumn{4}{c}{(b) DN, SwinIR, CBSD68} & \multicolumn{4}{c}{(c) DN, DMID, Urban100}  \\ \midrule
        
MUSIQ   & 70.945 & 70.727 & 70.722 & 65.713 & 70.105 & \underline{70.299} & 71.718 & 66.980 & 70.204 & 69.748 & 70.933 & 68.556 \\

MANIQA  & 0.4272 &  \textbf{0.4544}  &  0.4494  &  0.3947  & 0.4498 &         \underline{0.4770} &  0.5010 & 0.4309 & 0.4820 & 0.4784  & 0.5064 & 0.4613\\

Q-Align &  3.9245 &  \textbf{4.0529}  & 4.0134  & 3.6449  &  3.8287 &         \underline{4.1135} & 4.1346 & 3.6564 &  4.3794 & \underline{4.4836} & 4.5004 & 4.2940\\
NIQE $\downarrow$    &  4.9445 &  \underline{4.8870}  &  4.7881 & 4.5653 &  4.2245 &  \textbf{4.0804}      &  4.1256 & 3.9773 & 5.4216 & \underline{5.2342} & 5.1956 &  5.1961  \\
CLIPIQA & 0.4699 &  \underline{0.5264} & 0.5296 & 0.5157  & 0.6462 &  \underline{0.6759} & 0.6882 & 0.6459 & 0.5024 & \underline{0.5125} & 0.5477 & 0.5048 \\
LIQE    &  3.7250 &  \textbf{4.2317}  &  4.1447 &  3.9746  & 3.4197 & \textbf{4.2643}       & 4.2302 & 3.6651 & 3.9983 & \underline{4.2466} &  4.2707 & 4.1177       \\
\midrule \midrule
& \multicolumn{4}{c}{(d) CISR, HAT, Set5}    & \multicolumn{4}{c}{(e) RISR, OSEDiff, DRealSR} & \multicolumn{4}{c}{(f) RISR, OSEDiff, $\text{DIV2K-Val}^{\dagger}$}  \\ \midrule

MUSIQ   & 61.245 & 60.900 & 65.809 & 57.228 & 64.692 & \underline{65.358} & 66.011 & 60.882 & 67.934 & \underline{68.088} & 69.087 & 64.814 \\

MANIQA
& 0.4450 & \underline{0.4627}  &  0.5109 & 0.4168 & 0.4657 & \textbf{0.4837} & 0.4778     & 0.4220 & 0.5376 & \textbf{0.5605} & 0.5521 & 0.4754 \\

Q-Align & 3.0594 & \underline{3.1937} & 3.4683  & 2.8308  & 3.5433       & \underline{3.6231}  & 3.7180 & 3.2610 & 3.9313 & \underline{4.0938} &  4.1592  & 3.6756 \\
NIQE $\downarrow$    & 7.0979 & \underline{6.7614} & 6.4223 & 6.5715  & 6.4642  &  6.8326 & 6.2770 & 6.2491 & 6.0112 & 6.1792 & 5.7260  &  5.1601 \\
CLIPIQA &  0.5892 & \underline{0.6461} &  0.7140 & 0.6963 & 0.6347       & 0.6676 & 0.5787 &  0.4456 & 0.7916 & 0.7514 & 0.7682 & 0.7387 \\
LIQE    & 3.9618 &  \textbf{4.4055} & 4.2013 & 4.0362  & 3.9697       & 3.9015 &  4.0383 & 3.6189 & 4.3551 &  4.3004 & 4.4513 & 4.1684 \\
\bottomrule
\end{tabular}%
}
\caption{More quantitative comparisons across various IR datasets. We \underline{underline} the results where the optimized image $y_{\text{TTPO}}$ outperforms the restored image
$y_0$ in evaluation metrics. If $y_{\text{TTPO}}$ surpasses the preferred sample $y_w$, we further highlight the result in \textbf{bold}. Notably, we also provide numerical metric values for $y_w$ and $y_l$ in the Table as references. $^{\dagger}$For computational efficiency, we select the first 104 images (out of 3000 cropped patches) from the DIV2K-Val dataset.}
\label{tab:additional}
\end{table*}

\section{More Ablation Studies}
\label{sec:moreablation}

\noindent\textbf{Hyperparameters.} We provide ablation studies about the hyperparameters of TTPO, including the weighting parameter $\alpha$, the scaling factor $g$. As demonstrated in Tab.~\ref{tab:hyper}, the scaling factor $g$ significantly affects the performance, as it must appropriately scale the reward signals to align with the magnitude of the pre-trained denoising direction $\Delta t\cdot v_\theta(x_t,t)$. If $g$ is too large, it can distort the pre-trained denoising trajectory, resulting in an unstable optimization process. Conversely, if $g$ is too small, the guidance becomes insufficient to yield suboptimal optimization results. However, the value of $\alpha$ does not significantly affect the final results. Therefore, we simply select the value that yields the best performance as the choice for all the experiments in this paper.

\begin{table}[]
\centering

\resizebox{\linewidth}{!}{%
\begin{tabular}{@{}c|ccc||c|ccc@{}}
\toprule
$g$    & MUSIQ & MANIQA & Q-Align & $\alpha$   & MUSIQ & MANIQA & Q-Align \\ \midrule
3000 &   73.306    &   0.4730 &  4.0478  & 0.7 &  74.228 &  0.4886   &  4.3122  \\
2000 & 73.853 & 0.4869 & 4.1652 & 0.6 &  74.231 &  0.4909  &  4.3158       \\
1500 & 74.095 & \textbf{0.4918} &  4.2337  & 0.5 &  \textbf{74.242} & \textbf{0.4912}  & \textbf{4.3201}      \\
1000 & \textbf{74.242} & 0.4912  & \textbf{4.3201} & 0.4 & 74.213 &  0.4918      &     4.3157    \\
500  & 73.461 & 0.4725 & 4.2046 & 0.3 &  74.198 &  0.4899  &  4.3078       \\ \bottomrule
\end{tabular}%
}
\caption{Results of hyperparameters ablation studies. We report results on image denoising (DN), SwinIR~\cite{liang2021swinir}, Kodak24~\cite{kodak}.}
\label{tab:hyper}
\end{table}

\begin{table}[]
\centering

\resizebox{0.65\linewidth}{!}{%
\begin{tabular}{@{}c|ccc@{}}
\toprule
Steps    & MUSIQ & MANIQA & Q-Align \\ \midrule
28 &   71.410 &  0.4232 &  4.0443   \\
50 & 74.242 & 0.4912  & 4.3201  \\
100 & 74.251 & 0.4925 &  4.3254  \\ \bottomrule
\end{tabular}%
}
\caption{Ablations for denoising steps. We report results on image denoising (DN), SwinIR~\cite{liang2021swinir}, Kodak24~\cite{kodak}.}
\label{tab:steps}
\end{table}

\noindent\textbf{Number of Denoising Steps.} Recent pre-trained diffusion models possess the ability to generate high-quality images within a few denoising steps (\eg, 28 steps for FLUX). However, TTPO requires more denoising steps to incorporate sufficient guidance information for effective preference optimization. As shown in Tab.~\ref{tab:steps}, using only a few denoising steps results in suboptimal performance due to insufficient preference guidance, whereas increasing the number of steps leads to slight improvements at the cost of computational time.

\section{Limitations}
\label{sec:limi}
Our method has two main limitations. First, the inference process of FLUX requires significant computational resources, resulting in relatively long runtime, particularly when scaling to larger images or more complex datasets. To address this issue, we will explore more efficient methods for incorporating preference guidance in future work to improve computational efficiency. Second, although NR-IQA metrics show promising results, they still exhibit a noticeable gap when compared to human preferences (\eg, as evidenced in Fig.~\ref{fig:supp} and Tab.~\ref{tab:additional}, where some of the preferred samples show worse visual quality than the optimized image with better metric values), which makes it difficult for metric-based selection to fully align the optimization results with human judgment. Improving the computational efficiency of TTPO and developing more accurate and reliable preference selection mechanisms are important directions for our future work.

\bibliography{aaai2026}

\end{document}